%% file: main.tex
\documentclass[10pt,twocolumn,letterpaper]{article}

\usepackage{cvpr}              

\input{preamble}

\definecolor{cvprblue}{rgb}{0.21,0.49,0.74}
\usepackage[pagebackref,breaklinks,colorlinks,citecolor=cvprblue]{hyperref}

\def\paperID{4432} 
\def\confName{CVPR}
\def\confYear{2024}

\title{Instance-aware Exploration-Verification-Exploitation \\ for Instance ImageGoal Navigation}

\author{Xiaohan Lei$^{1}$ \quad Min Wang$^{2}$  \quad Wengang Zhou$^{1,2}$  \quad Li Li$^{1}$  \quad Houqiang Li $^{1,2}$ \thanks{This work was supported in part by the National Natural Science Foundation of China under Contract 62102128 and 62021001, and in part by the Fundamental Research Funds for the Central Universities under contract WK3490000007. It was also supported by the GPU cluster built by MCC Lab of Information Science and Technology Institution of USTC. Corresponding authors: Min Wang and Wengang Zhou.}\\
\normalsize $^{1}$MoE Key Laboratory of Brain-inspired Intelligent Perception and Cognition, Univerisity of Science and Technology of China\\
\normalsize $^{2}$Institute of Artificial Intelligence, Hefei Comprehensive National Science Center\\
{\tt\small leixh@mail.ustc.edu.cn, wangmin@iai.ustc.edu.cn, \{zhwg,lil1,lihq\}@ustc.edu.cn} \\
Project page: \url{https://xiaohanlei.github.io/projects/IEVE/}
}

\begin{document}
\maketitle

\begin{abstract}
   As a new embodied vision task, Instance ImageGoal Navigation (IIN) aims to navigate to a specified object depicted by a goal image in an unexplored environment. 
   The main challenge of this task lies in identifying the target object from different viewpoints while rejecting similar distractors. 
   Existing ImageGoal Navigation methods usually adopt the simple Exploration-Exploitation framework and ignore the identification of specific instance during navigation. 
   In this work, we propose to imitate the human behaviour of ``getting closer to confirm" when distinguishing objects from a distance. 
   Specifically, we design a new modular navigation framework named Instance-aware Exploration-Verification-Exploitation (IEVE) for instance-level image goal navigation. 
   Our method allows for active switching among the exploration, verification, and exploitation actions, thereby facilitating the agent in making reasonable decisions under different situations. 
   On the challenging HabitatMatterport 3D semantic (HM3D-SEM) dataset, our method surpasses previous state-of-the-art work, with a classical segmentation model (0.684 \textit{vs.} 0.561 success) or a robust model (0.702 \textit{vs.} 0.561 success).
\end{abstract}

\input{sections/introduction}

\input{sections/related_work}

\input{sections/methods}
\input{sections/experiment_setup}
\input{sections/conclusion}
\small\bibliographystyle{ieeenat_fullname}
\bibliography{main}


\end{document}

%% file: preamble.tex
\usepackage[dvipsnames]{xcolor}
\usepackage{multirow}


%% file: sections/introduction.tex
\vspace{-10pt}

\section{Introduction}
\label{sec:intro}

Embodied navigation is an emerging computer vision task, where an agent utilizes visual sensing to actively interact with its surroundings and execute navigation tasks~\cite{anderson2018evaluation, batra2020objectnav}. 
Over the past decade, substantial advancements have been witnessed in the field of embodied visual navigation, driven by the emergence of large-scale photorealistic 3D scene datasets \cite{xia2018gibson, straub2019replica, ramakrishnan2021habitat, chang2017matterport3d, yadav2023habitat}, the implementation of advanced simulators \cite{kolve2017ai2, savva2019habitat, xia2018gibson, szot2022habitat}, the development of deep learning algorithms \cite{schulman2017proximal, he2022masked, caron2021emerging, oquab2023dinov2}, and so on.

\begin{figure}
  \centering
  \includegraphics[width=0.99\linewidth]{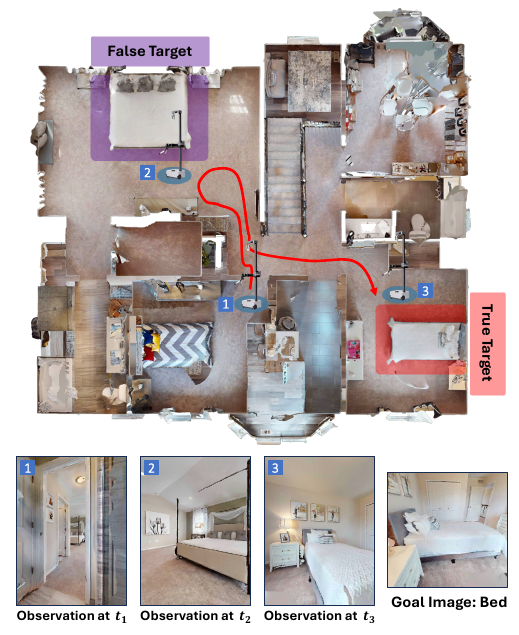}
    \vspace{-6pt}
  \caption{{\bf Instance ImageGoal Navigation} tasks an agent with navigating to a particular object instance described by the goal image. The agent is initially in an unexplored environment at $t_1$. While exploring the environment, it encounters a ``bed" similar to the target at $t_2$ and discerns their differences. Eventually, at $t_3$, agent finds the ``bed" described in the goal image.}
  \label{fig:intro}
  \vspace{-20pt}
\end{figure}

The ability for an agent to navigate its environment and accomplish tasks without human intervention is essential for the advancement of embodied visual navigation. 
One common setting for tasking an agent is to give it natural language instructions, such as ``check if my laptop is on the chair." 
However, this setting becomes confusing when there are multiple chairs in the house. 
To overcome this challenge, Krantz~\etal \cite{krantz2022instance} propose the Instance ImageGoal Navigation task (IIN), which involves providing the agent with an image captured from goal camera that depicts a specific instance. 
The agent is then required to navigate to the location of the instance depicted in the image, as shown in \cref{fig:intro}. By utilizing the visual information, the agent is expected to avoid confusion caused by multiple similar distractors of the same class.

The most relevant task with IIN is ImageGoal Navigation \cite{al2022zero, Chaplot_2020_CVPR, hahn2021no, majumdar2022zson, mezghan2022memory, yadav2023offline, zhu2017target}, which has been a popular topic in embodied visual navigation research. The difference between IIN and ImageGoal Navigation lies in their requirements for navigation success. 
In other words, IIN demands that the agent has the ability to recognize the same object from different viewpoints, while ImageGoal Navigation requires maximizing the similarity between the agent's observations and the goal image. 
Previous ImageGoal navigation requires targeting random goal images captured by cameras that share the same parameters like height, look-at-angle, and field-of-view as the agent camera. 
Some approaches \cite{wasserman2022lastmile, krantz2023navigating} employ local feature matching algorithms and make binary decisions based on the current observation. 
They typically expect the agent to accurately determine whether the current observation matches the goal image. 
However, it is extremely challenging to enable the agent to correctly identify the target object from a considerable distance. In addition, they treat the decision-making process as a one-shot task, perceiving it as a match between the current observation and the target, while overlooking the judgment process across the temporal dimension.



To address the above limitations, we propose a novel IIN framework, \emph{i.e.,} Instance-aware Exploration-Verification-Exploitation (IEVE), for Instance ImageGoal Navigation. 
Inspired by the human behavior of ``getting closer to confirm" when recognizing distant objects, we formulate the task of determining whether an object matches the one in the goal image as a sequential decision problem. 
In addition, we design a novel matching function that relies not only on the current observation and goal image but also on the Euclidean distance between the agent and the potential target. 
We categorize the targets into confirmed target, potential target, and no-target (exploration), and allow the agent to actively choose among these three targets. 
After an efficient training process of our framework, our result on the IIN task outperforms previous state-of-the-art methods, achieving $0.684$ \textit{Success} with a classical instance segmentation model \cite{he2017mask}, and $0.702$ with a robust model \cite{wang2023internimage}.

%% file: sections/related_work.tex
\section{Related Work}

\begin{figure*}
  \centering
  \includegraphics[width=0.99\linewidth]{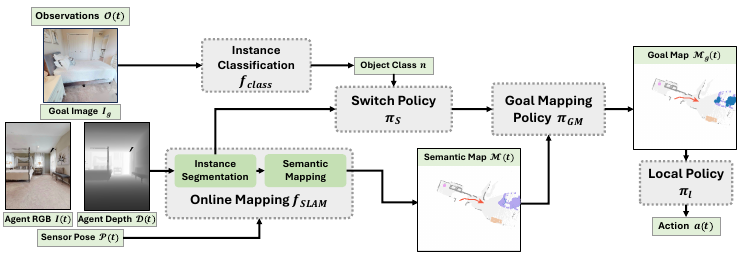}
  \caption{{\bf Framework Overview.} Our model consists of five main components. Instance Classification $f_{class}$ predicts the object's class in goal image $n$. Online Mapping $f_{SLAM}$ uses RGB-D and sensor pose reading $\mathcal{P}(t)$ to construct a semantic map $\mathcal{M}(t)$ of the environment. The Switch Policy $\pi _S$ and Goal Mapping Policy $\pi _{GM}$ are interconnected. The Switch Policy $\pi _S$ determines the output of the Goal Mapping Policy's goal map $\mathcal{M}_g(t)$ based on its input judgments (whether a potential target exists and whether the potential target is confirmed). Once the goal map $\mathcal{M}_g(t)$ is determined, the Local Policy $\pi _l$ is used to determine the action $a(t)$ taken by the agent at the current timestep.}
  \label{fig:main}
  \vspace{-10pt}
\end{figure*}

In recent years, there has been great research interest in the field of embodied vision and active agents. Embodied agents are required to effectively and creatively resolve complex tasks and interact with the physical environment. Notably, the navigation task plays a pivotal role in embodied vision. 
Numerous researchers have devoted their efforts to exploring different forms of navigation problems, encompassing various task goals and input formats. 
Specifically, there are five main categories of visual navigation problems: Point Goal Navigation \cite{anderson2018evaluation, savva2017minos, chaplot2020learning} 
, Object Goal Navigation \cite{batra2020objectnav, savva2017minos, Gadre_2023_CVPR, Du_2023_CVPR, Zhang_2023_CVPR}, 
, Visual Language Navigation \cite{anderson2018vision, krantz2020beyond, Krantz_2023_CVPR, Hwang_2023_CVPR, Yang_2023_CVPR, Gao_2023_CVPR, Huo_2023_CVPR, Kamath_2023_CVPR, Li_2023_CVPR,Li_2023_CVPR}, 
ImageGoal Navigation 
 \cite{al2022zero, Chaplot_2020_CVPR, hahn2021no, majumdar2022zson, mezghan2022memory, yadav2023offline, zhu2017target, Kwon_2023_CVPR},  
 and Instance ImageGoal Navigation \cite{krantz2023navigating}.
 According to the learning framework design, existing visual navigation methods are typically divided into two categories: modular and end-to-end approaches.


End-to-end navigation policies are commonly optimized with improved visual representations \cite{mousavian2019visual, yang2018visual}, auxiliary tasks \cite{ye2021auxiliary}, carefully reward-shaping \cite{maksymets2021thda} and data augmentation techniques \cite{maksymets2021thda}. 
Visual representations can be improved with fine-tuned segmentation model \cite{mousavian2019visual} and object relation graphs \cite{druon2020visual, du2020learning, moghaddam2021optimistic, pal2021learning, zhang2021hierarchical}.
Prior works also learn fine-tuning visual encoder with Self-Supervised Learning (SSL) \cite{partsey2022mapping, yadav2023ovrl}, while some researchers employ imitation learning approaches, such as Ramrakhya \etal \cite{ramrakhya2022habitat}, gathering data from large-scale human annotations. 
Although achieving promising navigation success in simulator, end-to-end methods often suffer high computation complexity in training and generalizability, which may degrade the success of navigation in diverse environments. 

To address these challenges, modular methods have been proposed to solve navigation tasks with separate modules, including perception, mapping, path planning, \textit{etc}. Chaplot \etal \cite{chaplot2020object} address the Object Goal Navigation problem using semantic mapping and exploration. Majumdar \etal \cite{majumdar2022zson} adopt the zero-shot learning and inference approach with the use of CLIP. Rmakrishnan \etal \cite{ramakrishnan2022poni} propose a potential field function to tackle the exploration problem in Object Goal Navigation. Mezghan \etal \cite{mezghan2022memory} and Savinov \etal \cite{savinov2018semi} construct augmented memory mechanisms in ImageGoal Navigation. Some approaches \cite{wasserman2022lastmile, krantz2023navigating} employ local feature matching algorithms and make binary decisions based on the current observation for ImageGoal Navigation. In general, modular-based methods typically exhibit excellent scalability with lower training cost.

In this work, we focus on IIN task, which is a newly formulated but more real navigation task. It is relevant with the classical ImageGoal Navigation task. ImageGoal Navigation is commonly studied in a previous-unseen environment where goal images are randomly sampled sharing the same parameter with the agent's camera \cite{zhu2017target}. Many approaches study the ImageGoal Navigation task with RL to learn policies that directly map observations to action \cite{al2022zero, yadav2023offline, yadav2023ovrl, choi2021image}. Yadav \etal \cite{yadav2023ovrl} postulate that SSL for visual representation greatly contributes to the success of the end-to-end navigation policy training. Choi \etal \cite{choi2021image} attribute the success of RL training to the carefully reward shaping. Chaplot \etal \cite{Chaplot_2020_CVPR} propose a topological mapping system to address the ImageGoal navigation problem, as well as Kim \etal \cite{TSGM}. Kwon \etal \cite{Kwon_2023_CVPR} propose a novel type of map for ImageGoal Navigation, a renderable neural radiance map. 

Unlike ImageGoal Navigation, IIN \cite{krantz2022instance} focuses on distinguishing different similar instances and being able to identify the object in the goal image at different viewpoints. On the other hand, ImageGoal Navigation pays more attention to maximizing the similarity between observations and the goal image within an episode. To be more specific, the goal camera in IIN task is disentangled from the agent’s camera on the sampled parameters such as height, look-at-angle, and field-of-view reflect the realistic use case of a user-supplied goal image. Additionally, previous state-of-the-art work employs the confidence score for object re-identification. Differently, we propose a new switchable strategy to meticulously examine a currently visible object. Our key insight lies in treating the subtask of target confirmation as a sequential decision problem and designing a new matching function, which enables the agent to more accurately distinguish target objects from misleading ones.


\begin{figure*}
\centering
\includegraphics[width=0.99\linewidth]{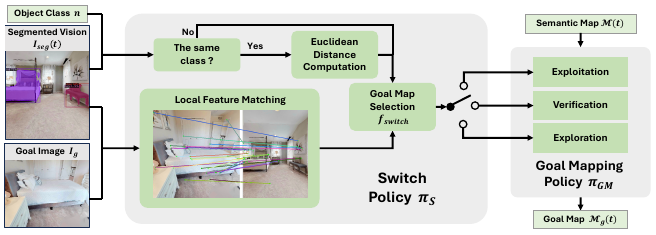}
\caption{{\bf Switch Policy and Goal Mapping Policy.} The Goal Map Selection module $f_{switch}$ receives two inputs: the shortest Euclidean distance between the agent and the potential target (or non-existence), and the number of matched keypoints. These inputs are mapped to a selection signal using Goal Map Selection function $f_{switch}$. At each timestep, the Switch Policy $\pi _S$ will choose one and only one of the three parallel modules of Goal Mapping Policy $\pi _{GM}$.}
\label{fig:policy}
  \vspace{-10pt}
\end{figure*}

%% file: sections/methods.tex

\section{Method}

We devise a new framework, Instance-aware Exploration-Verification-Exploitation (\texttt{IEVE}) for Instance ImageGoal Navigation, which consists of five key modules, including Instance Classification, Online Mapping, Switch Policy, Goal Mapping Policy and Local Policy, as depicted in \cref{fig:main}. 
In the following subsections, we will first introduce the preliminaries in \cref{sec:preli}, and sequentially elaborate the five modules of our framework.

\subsection{Preliminaries} \label{sec:preli}

We follow the embodiment setting of Hello Robot Stretch \footnote{\url{https://hello-robot.com/stretch-2}}. The agent in the simulation environment is a rigid-body, zero-turn-radius cylinder of height 1.41m and radius 0.17m. The forward-facing RGB-D camera is mounted at a height of 1.31m. At each timestep $t$, the agent's observation $\mathcal{O}_t$ consists of an egocentric RGB image $\mathcal{I}_t$, depth image $\mathcal{D}(t)$, goal image $\mathcal{I}_g$ and sensor pose reading $\mathcal{P}(t)$. Camera parameters of $\mathcal{I}_t$ and $\mathcal{I}_g$ like height, look-at-angle and field-of-view are not the same. Specifically, $\mathcal{I}_t$ is a $640\times480$ image while $\mathcal{I}_g$ is a $512\times512$ image.

\subsection{Instance Classification}

 At the onset of each episode, the Instance Classification model will first process $\mathcal{I}_g$ and output an object class label $n$. We adopt two different types of models to achieve instance classification, including 1) classification model:  we use the goal images on the train split of HM3D-SEM \cite{yadav2023habitat} to finetune the image classification model, \textit{i.e.}, Swin Transformer \cite{liu2021swin}, pretrained on ImageNet \cite{5206848}. It may face the inherent semantic ambiguity of Ground Truth labels in the HM3D-SEM, such as the difficulties in distinguishing between the class ``sofa" and ``chair". 2) instance segmentation model: we use the semantic label of the most central segmented instance in the goal image as the classification result. This approach can maintain semantic consistency without being affected by Ground Truth labels.

\subsection{Online Mapping}


At each time step $t$, the agent will utilize the egocentric view to construct the current semantic map $\hat{m}(t)$ within its local coordinate system. The RGB frame $\mathcal{I}(t)$ is first processed through instance segmentation (Mask RCNN \cite{he2017mask}), yielding the output $\mathcal{I}_{seg}(t)$ with dimensions of $(B, N+3, H, W)$, denoting batch size, the number of classes $N$ plus RGB channels of $\mathcal{I}(t)$, frame height, and frame width. Simultaneously, the depth image $\mathcal{D}(t)$ is transformed into a pointcloud $P_c(t)$ using the camera's intrinsic matrix $\boldsymbol{T}_{int}$, followed by the voxelization of the pointcloud. Subsequently, the feature points of $\mathcal{I}_{seg}(t)$ are scattered into the voxels, resulting in the overall local map construction process being formulated as:
\begin{equation}
\hat{m}(t) = \text{Scatter}(\text{Voxel}(\boldsymbol{T}_{int}\mathcal{D}(t)), \mathcal{I}_{seg}(t)).
\label{eq:semantic_mapping}
\end{equation}

With the constructed semantic map $\hat{m}(t)$ within local coordinate system, agent will subsequently conduct max pooling with the previous global semantic map to obtain the global semantic map $\mathcal{M}(t)$ at timestep $t$ as follows:
\begin{equation}
\mathcal{M}(t) = \text{MaxPooling}(\mathcal{M}(t-1), \boldsymbol{T}_{lg}\hat{m}(t)),
\label{eq:maxpooling}
\end{equation}
where $\boldsymbol{T}_{lg}$ denotes the transformation matrix from the agent's local frame to the world's global coordinate system. The size of $\mathcal{M}(t)$ is specified as $(B, 2+N, H_m, W_m)$, where the first channel represents the explored area of the entire environment, the second channel defines the non-traversible area, and the last $N$ channels signify the location of each object class. 


\subsection{Switch Policy}

At each time step $t$, Switch Policy actively selects the functional goal map from the three parallel sub-modules in Goal Mapping Policy. We summarize the switch process as a function with segmented depth image $\mathcal{I}_{seg}(t, n)\cdot\mathcal{D}(t)$, the number of matched keypoints $\omega$ using local feature matching algorithm and the minimum Euclidean distance $d$ between the current position of the agent and the potential target (if exists). This can be described as follows:
\begin{equation}
S = f_{switch}(\mathcal{I}_{seg}(t, n)\cdot\mathcal{D}(t), \omega, d),
\label{eq:switch}
\end{equation}
where $S$ indicates which goal mapping policy to be selected, depicted in \cref{fig:policy}. To obtain the specific expression of $f_{switch}$, we construct an Instance Re-Identification dataset as depicted in \cref{fig:dataset} to illuminate the influence of distance $d$ and the number of matched keypoints $w$ on the goal object's re-identification. 

\begin{figure}[t]
  \centering
  \includegraphics[width=0.90\linewidth]{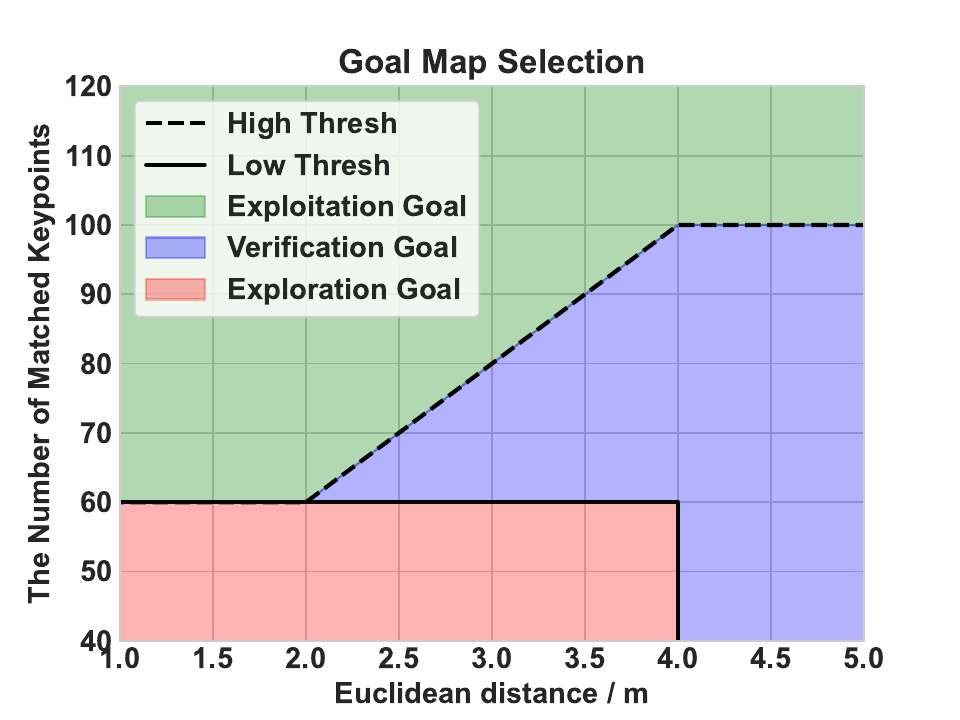}
  \vspace{-6pt}
  \caption{Goal Map Selection function $f_{switch}$ with respect to the Euclidean distance from agent to the potential target and the number of matched keypoints. Best viewed in color.}
  \label{fig:selection}
  \vspace{-6pt}
\end{figure}

The Goal Map Selection function $f_{switch}$ aligns with human experience in object verification. $f_{switch}$ sets objects of the same category as potential targets. At a larger distance to make accurate decisions, a higher threshold on number of matched keypoints is set to convert a potential target into a determined target as depicted in \cref{fig:selection}. As the distance decreases, the upper threshold also degrades continuously. At closer distance, the agent's judgment of the object becomes more reliable, so when the number of matched keypoints falls below the lower threshold, it can be very reliably determined that the potential target is not the goal target, thereby switching to the Exploration Policy.

{ \bf Local Feature Matching.} We employ a keypoint-based re-identification method to help in the selection of goal mapping policy in \cref{eq:switch}. Conditioned on the goal image $\mathcal{I}_g$ and egocentric image $\mathcal{I}(t)$, we extract the pixel-wise $(x, y)$ coordinates of the keypoints $K_t \in \mathbb{R} ^ {n \times 2}$ and their associated feature descriptors $V_t \in \mathbb{R} ^ {n \times 256}$. These features are extracted utilizing DISK \cite{tyszkiewicz2020disk}, an end-to-end local feature learning algorithm that utilizes policy gradient. 
Likewise, the same process is repeated with the goal image, resulting in $K_g \in \mathbb{R} ^ {m \times 2}$ and $V_g \in \mathbb{R} ^ {m \times 256}$. 
The feature extraction processes are formulated as follows:
\begin{equation}
(K_t, V_t) = \text{DISK}(\mathcal{I}(t)),  \quad (K_g, V_g) = \text{DISK}(\mathcal{I}_g).
\label{eq:feature1}
\end{equation}
Subsequently, the matched pairs $(\hat{K_t}, \hat{K_g}) \in \mathbb{R} ^ {\omega \times 2}$ are computed using LightGlue \cite{lindenberger2023lightglue}, a deep neural network that matches sparse local features across image pairs. The feature matching process is formulated as follows: 
\begin{equation}
(\hat{K_t}, \hat{K_g}) = \text{LightGlue}((K_t, V_t), (K_g, V_g)).
\label{eq:matching}
\end{equation}
Subsequently, the number of matched points $\omega$ is utilized as an adaptive threshold for the verification and exploitation goal selection.

\subsection{Goal Mapping Policy}

Depicted in \cref{fig:policy}, Goal Mapping Policy module is divided into three parallel sub-modules, which will function at different timesteps. In other words, at each timestep, only one sub-module will function. More specifically, given the predicted goal object's class $n$ and segmented vision $\mathcal{I}_{seg}(t)$, if the mask $\mathcal{I}_{seg}(t, n)$ has any value greater than zero, then the Exploitation or Verification Policy is activated. They project the masked depth image $\mathcal{I}_{seg}(t, n)\cdot\mathcal{D}(t)$ to current semantic map. Similar to previous Online Mapping module, the two goal mapping policies in local coordinate system can be summarized as follows:
\begin{equation}
\hat{m}_g(t) = \text{Voxel}(\boldsymbol{T}_{int}\mathcal{I}_{seg}(t, n)\cdot\mathcal{D}(t)),
\label{eq:goal_localization}
\end{equation}
where $\hat{m}_g(t)$ indicates the goal map in the local coordinate system. Then we transform $\hat{m}_g(t)$ into the global coordinate system, $\mathcal{M}_g(t)=\boldsymbol{T}_{lg}\hat{m}_g(t)$. Although with the same goal map formulas, Exploitation and Verification policy represent the targets with different confidence. Specifically, Exploitation policy refers to the confirmed target, which will be kept and used for the rest navigation process. In contrast, Verification policy represents potential target, which will be further checked by local feature matching in the Switch policy module.

The Exploration Policy use a convolutional neural network to predict a waypoint $g_{rl}(t)=\pi _{rl}(\mathcal{M}(t)|\theta_{rl})$, where $\theta_{rl}$ are parameters of Exploration Policy. We train it with RL to maximize coverage of the environment while also successfully employing the \textit{stop} action upon encountering a correct oracle facing the goal object. The goal map $\mathcal{M}_g(t)$ is constructed with the value at position $g_{rl}(t)$ set as 1.


\input{tables/main_table}

\subsection{Local Policy}
When Switch Policy selects one of three goal mapping policies, agent adopts Local Policy to infer next action in accordance with the semantic map $\mathcal{M}(t)$, and the goal map, $\mathcal{M}_{g}(t)$. Specifically, Fast Matching Method (FMM) is employed to generate a distance field, which establishes the shortest distance between a traversible point and the selected goal map. A subset of the distance field is chosen within the agent's feasible range, and a waypoint is selected, ensuring that it does not intersect with existing obstacles and maintains the local minimum within the subset. With the selected waypoint, the agent can readily calculate an action based on the angle and distance from its current state. We also add a visited trajectory mechanism for the agent to avoid been stuck by invisible obstacles, which will be discussed in the supplementary materials. In our framework, only Instance Classification and Exploration Policy require few training, underscoring the computation efficiency of our design.


%% file: tables/main_table.tex
\begin{table}
  \centering
  \begin{tabular}{ccc}
    \toprule
    & \multicolumn{2}{c}{\textbf{Validation}} \\
    \cmidrule{2-3}
    Model & \textit{Success}~$\uparrow$ & \textit{SPL}~$\uparrow$ \\
    \midrule
    RL Baseline \cite{krantz2022instance} & 0.083 & 0.035  \\
    OVRL-v2 ImageNav \cite{yadav2023ovrl} & 0.006 & 0.002 \\
    OVRL-v2 IIN \cite{yadav2023ovrl}* & 0.248 & 0.118 \\
    Mod-IIN \cite{krantz2023navigating} & 0.561 & 0.233 \\
    \midrule
    IEVE Mask RCNN (ours) & 0.684 & 0.241 \\
    \textbf{IEVE InternImage (ours)} & \textbf{0.702} & \textbf{0.252} \\
    \bottomrule
  \end{tabular}
  \caption{Comparison between our IEVE and prior methods on IIN task. ``*" means we finetune OVRL-v2~\cite{yadav2023ovrl} on the IIN task. \texttt{IEVE} outperforms the baseline model (row 1) with up to $8.75$x increase in \textit{Success} and outperforms a state-of-the-art model \cite{krantz2023navigating} (Mod-IIN, row 4) from $0.561$ to $0.684$ with a lightweight segmentation model (Mask RCNN) on \textit{Success}, and higher \textit{Success} of $0.702$ with a strong backbone (InternImage \cite{wang2023internimage}).}
  \label{tab:main}
  \vspace{-10pt}
\end{table}

%% file: sections/experiment_setup.tex
\section{Experiment}

\subsection{Experiment Setup}

{\bf Dataset.} For our experiment, we employ the Habitat \cite{szot2022habitat} simulator to conduct our experiments. The definition of the Instance ImageGoal Navigation \cite{krantz2022instance} task utilized in the Habitat Navigation Challenge 2023 \footnote{\url{https://aihabitat.org/challenge/2023/}} serves as our primary reference point. The episode dataset proposed by Krantz et al. \cite{krantz2023navigating} for Instance ImageGoal Navigation serves as our key data source. The scene dataset utilized in this study stems from the Habitat-Matterport3D with semantic annotations (HM3D-SEM) \cite{yadav2022habitat}, which encompasses a total of 216 scenes. 
These scenes are split into three distinct subsets for training, validation, and testing purposes, consisting of 145/36/35 scenes, respectively. 
Additionally, the episode dataset has been partitioned into four subsets for training, validation, testing, and testing under a modified challenge regime, comprising 7,056K/1K/1K/1K episodes respectively. 
We train our agent on the train split and evaluate it on validation split. 
The object depicted by the goal image belongs to the following six categories:  $\{$ \textit{chair, couch, plant, bed, toilet, television} $\}$. 
On the validation subset, a total of 795 unique object instances have been observed.

{\bf Action Space.} Adhering to the task setting of the Habitat Navigation Challenge 2023, we employ the continuous action space. The action space encompasses four actions: \textit{linear velocity, angular velocity, camera pitch velocity, velocity stop}, each of which accepts values ranging between $-1$ and $1$ and is appropriately scaled according to the respective configurations. Our linear speed is capped at a maximum of $35cm/frame$ and angular velocity at $60^\circ/frame$.

{\bf Metrics.} We evaluate our model with both success and efficiency. We report Success Rate (\textit{Success}), Success rate weighted by normalized inverse Path Length (\textit{SPL}). \textit{Success} is True if the agent calls \textit{velocity stop} action within $1.0m$ Euclidean distance of the goal object and the object is oracle-visible by turning or looking up and down. \textit{SPL} is an efficiency measure defined in \cite{anderson2018evaluation}.

\begin{figure*}
  \centering
  \includegraphics[width=0.99\linewidth]{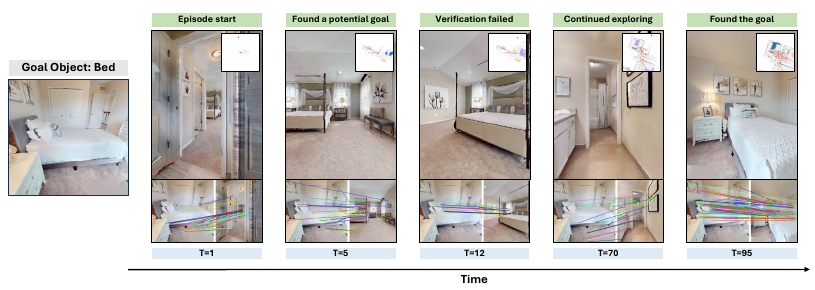}
  \vspace{-10pt}
  \caption{Qualitative example of our \texttt{IEVE} agent performing the Instance ImageGoal Navigation task in the Habitat simulator. Agent is initialized at $T=1$ , and finds a potential target at $T=5$. After carefully evaluating the potential target from $T=5$ to $T=12$, the agent proceeds with its exploration. Finally, agents identifies the goal bed instance at $T=95$.}
  \label{fig:time}
  \vspace{-10pt}
\end{figure*}

\subsection{Comparison with State-of-the-art Methods}
 We undertake a comprehensive evaluation of our proposed model against various baselines and the SOTA work.

{\bf End-to-End RL Agent.} The baseline model for the IIN task is trained in a fully-fledged end-to-end fashion. This involves inputting both an egocentric rgb and depth image as well as a pose into the model, which is then processed by a visual encoder and a recurrent neural network. The network output directly determines the agent's action. The agent is trained from scratch utilizing Proximal Policy Optimization (PPO) \cite{schulman2017proximal}. Our method exhibits 8.24x higher \textit{Success} in Val (row 5 \textit{vs.} row 1 in \cref{tab:main}). Although it takes a staggering 3.5 billion steps for RL agent model to achieve convergence, the generalization capability of the agent remains unsatisfactory.

{\bf OVRL-v2.} Offline Visual Representation Learning v2 (OVRL-v2 \cite{yadav2023ovrl}) was proposed by Karmesh \etal, who introduced a novel model-free, end-to-end navigation policy that emphasizes the significance of self-supervised pretraining of visual encoders and introduces the concept of Compression Layer (CL). Employing a ViT + LSTM backbone, this policy exhibits state-of-the-art performance in ImageGoal Navigation and approaches the state-of-the-art in ObjectGoal Navigation. Due to the task disparity between ImageGoal Navigation and Instance ImageGoal Navigation, we fine-tune the pretrained policy for our specific task.

The original OVRL-v2 is trained under the setting of the ImageGoal Navigation task. Evaluating OVRL-v2's performance without finetuning for Instance ImageGoal Navigation reveals suboptimal results (0.006 \textit{Success} row 2 in \cref{tab:main}). Several factors impact the performance: a shift in scene datasets (Gibson \cite{xia2018gibson} \textit{vs.} HM3D-SEM \cite{yadav2023habitat}), changes in embodiment (Locobot \textit{vs.} Stretch), and disparity in goal destinations (image source \textit{vs.} image subject). When specifically finetuned for Instance ImageGoal Navigation (row 3 in \cref{tab:main}) on HM3D-SEM, OVRL-v2 yields improved performance rates at 0.248 \textit{Success}. Our approach with Mask RCNN exhibits $0.436$ higher \textit{Success} in Val (row 5 \textit{vs.} row 3 in \cref{tab:main}), and $0.454$ higher \textit{Success} with InternImage (row 6 \textit{vs.} row 3 in \cref{tab:main}).

{\bf Mod-IIN.} Mod-IIN \cite{krantz2023navigating}
decomposes Instance ImageGoal Navigation task into exploration, goal re-identification, goal localization and local navigation. It outperforms previous state-of-the-art end-to-end learned policies without fine-tuning. Mod-IIN achieves $0.561$ \textit{Success} and $0.233$ \textit{SPL} (row 4 in \cref{tab:main}), and we surpass their results in \textit{Success} with a substantial margin of $0.123$ by attaining $0.684$ \textit{Success} and a marginal improvement in \textit{SPL} with the help of a classical visual backbone trained with Mask RCNN \cite{he2017mask} (row 5 in \cref{tab:main}). Additionally, with the current state-of-the-art Semantic Segmentation model InternImage \cite{wang2023internimage}, our results (row 6 in \cref{tab:main}) are notably superior with an impressive score of $0.702$ \textit{Success} and $0.252$ \textit{SPL}.

\input{tables/t_exp}

\subsection{Ablation Study}

\begin{figure}[t]
  \centering
   \includegraphics[width=1.0\linewidth]{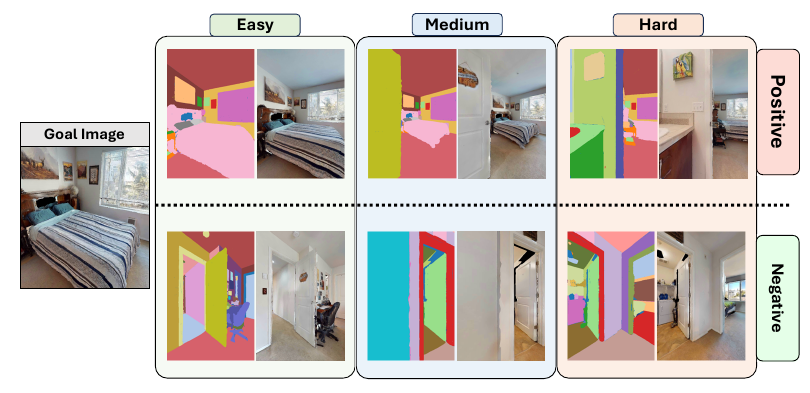}
   \caption{{\bf Instance Re-Identification Dataset.} Based on the object depicted in the goal image, we sample positive and negative samples with varying Euclidean distance which represents the difficulty level of instance re-identification. For each difficulty level, the left side shows the ground truth segmentation results and the right side shows the RGB images.}
   \label{fig:dataset}
   \vspace{-16pt}
\end{figure}

\begin{figure*}
  \centering
  \vspace{-20pt}
  \includegraphics[width=0.99\linewidth]{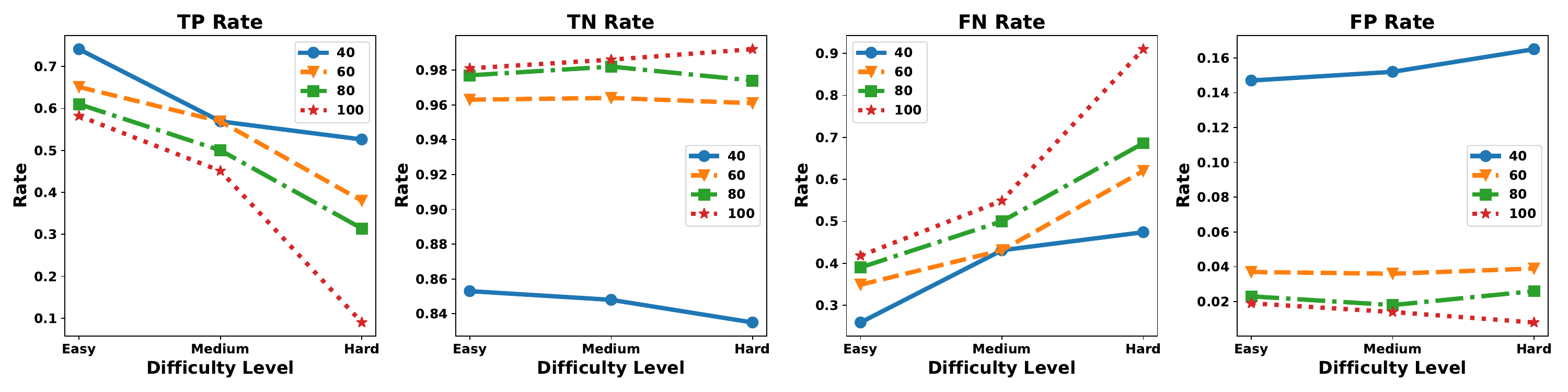}
  \caption{The TP, TN, FN and FP rates vary with changes in the threshold and difficulty levels (representing the closest Euclidean distance between the agent and the potential target). Threshold represents the threshold for the number of matched keypoints that determines whether two images are considered to be the same instance. TP decays rapidly as the difficulty increases, indicating that it becomes increasingly challenging for the agent to make correct judgments at a larger distance.}
  \label{fig:confusion}
  \vspace{-10pt}
\end{figure*}

\indent {\bf Instance Classification.} Our instance classification method falls into two categories. One is to fine-tune the pretrained SwinTransformer \cite{liu2021swin} on the goal images of the HM3D-SEM training set. The other method is to obtain the semantic category of the most central segmentation result of the goal image using the segmentation model. Ablation experiments conducted on these two methods demonstrate that SwinTrans is helpful for Mask RCNN (row 6 \textit{vs.} row 1 in \cref{tab:ablations}), but it actually suppresses the performance of InternImage (row 10 \textit{vs.} row 5 in \cref{tab:ablations}). We believe this is because the InternImage segmentation model has a larger parameter size, better zero-shot capabilities on this dataset, and stronger semantic labeling capabilities.

{\bf Exploration Policy.} To scrutinize the extent to which exploration policy exerts a meaningful influence on final outcomes, we separately train our framework with three different exploration policies: a random exploration policy, an RL exploration policy, and a frontier exploration policy. Each of these three policies generates an exploration goal map every ten local steps. The first random policy involves taking the semantic map as input and randomly generating a goal map. The second policy, trained through Reinforcement Learning, seeks to maximize coverage of the environment while also encouraging the \textit{velocity stop} action upon encountering a correct oracle facing the goal object. The frontier exploration policy, in contrast, endeavors to delineate the boundaries of the currently explored and free regions of the map, thereby constructing a goal map that encourages exploration. The implementation details of exploration policy are discussed in the supplementary materials.

From \cref{tab:ablations}, the performance of RL (row 1 and row 6 in \cref{tab:ablations}) and random (row 3 and row 8 in \cref{tab:ablations}) exploration policy on the \textit{SPL} significantly exceeds that of Frontier exploration policy (row 2 and row 7 in \cref{tab:ablations}). 
This disparity in performance is due to the fact that Frontier Exploration requires exploring every corner of a given environment, which is time-consuming and unnecessary in most cases. 
Specifically, the agent should focus on gaining a comprehensive understanding of the most salient aspects of a small room, which facilitates the identification of the target object.

{\bf Instance Segmentation.} Initially, we employ the Mask RCNN instance segmentation model on the Online Mapping module. In order to explore the influence of the visual model on the framework, we have employed the current state-of-the-art semantic segmentation approach: InternImage \cite{wang2023internimage} finetuned on the COCO dataset. Furthermore, in order to ascertain the upper bound of our method, we have employed the ground truth labeled instance class and the ground truth instance segmentation to conduct comprehensive evaluations of the \textit{Success} and \textit{SPL} rates. This entails the use of a local policy for the purpose of goal map-to-action, where the agent is directly capable of ascertaining whether the goal object is present in the current view while exploring the environment.

From \cref{tab:ablations}, we can observe that state-of-the-art semantic segmentation model \texttt{InternImage} (row 5 in \cref{tab:ablations}) indeed benefits navigation, and the ground truth segmented observation (last row in \cref{tab:ablations}) gives the upper bound of navigation. Specifically, more precisely segmented observation from \texttt{InternImage} gives higher metrics both in \textit{Success} and \textit{SPL}. This is due to the more precise goal localization and less wrong segmentation results.

{\bf Switch Policy.} We conduct a comparative analysis of Switch policy. We innovatively incorporate a verification policy into the decision-making process of the agent, and through ablation experiments (row 1 \textit{vs.} row 4 and row 6 \textit{vs.} row 9 in \cref{tab:ablations}), we demonstrate its effectiveness and significance. The implementation details of the Exploration-Exploitation Policy will be presented in supplementary materials. As depicted in \cref{tab:ablations}, our EVE policy (row 1, row 6 in \cref{tab:ablations}) outperforms Exploration-Exploitation (EE) Policy (row 4, row 9 in \cref{tab:ablations}) by a substantial margin both on \textit{Success} and \textit{SPL}. This suggests that the EVE policy provides the agent with greater opportunities to discern the current observation from the goal image, rather than arbitrarily assigning a false label to the current observed object in the EE policy. Depicted in \cref{fig:time}, agent finds the potential target at $T=5$ and employ verification policy. After carefully examining the similar object, the agent determines that the object is not the target and actively switches back to Exploration Policy.

{\bf Instance Re-Identification. \label{sec:instance}} To assess our Switch module and identify optimal hyperparameters of the Goal Map Selection module $f_{switch}$, we have constructed a comprehensive dataset, as depicted in \cref{fig:dataset}. This dataset is created based on the HM3D-SEM train/val split. Anchor images are defined as goal images that depict specific object instances, which do not share the same parameters as agent's camera.

For each anchor goal image, we sample 10 positive and negative images using the agent's camera. Positive images are defined as observations where the target instance is visible to the oracle, while negative images are randomly sampled from the same scene and do not contain the specified goal instance. We have divided the dataset into three distinct difficulty levels: {\textit{easy, medium, hard}}, as shown in \cref{fig:dataset}. Different difficulty levels represent different Euclidean distances from the target to current viewpoint. The further implementation details will be discussed in the supplementary materials. To demonstrate the capabilities of our Switch module, we employ the train split to determine hyperparameters and subsequently evaluate performance on the val split. Notably, the train and val splits are two distinct sets, which ensure that the agent has never encountered any scene or object instance present in the train split.



 As depicted in \cref{fig:confusion}, it is evident that with a fixed threshold of $60$, the TP values for the difficulty level decrease significantly. It becomes more evident in the hard level with a threshold of $100$ that yields TP rate close to $0$. 
 This highlights the importance of careful examination and decision-making based on the utilization of local matching feature algorithms to differentiate between similar objects. 
 By meticulously examining the possible object instance, the agent is able to confidently arrive at a decision. 
 We believe this explains why our model significantly surpasses previous state-of-the-art research in terms of \textit{Success}.

%% file: tables/t_exp.tex
\begin{table}[t]
  \centering
  \resizebox{\linewidth}{!}{
  \begin{tabular}{cccccc}
    \toprule
    \multicolumn{4}{c}{\textbf{Model Variation}} & \multicolumn{2}{c}{\textbf{Validation}} \\
    \cmidrule{5-6}
    Classification & Perception & Switch & Exploration & \textit{Success}~$\uparrow$ & \textit{SPL}~$\uparrow$ \\
    \midrule
    Mask RCNN & Mask RCNN & EVE & RL & 0.666 & 0.235  \\
    Mask RCNN & Mask RCNN & EVE & Frontier & 0.676 & 0.182 \\
    Mask RCNN & Mask RCNN & EVE & Random & 0.654 & 0.233 \\
    Mask RCNN & Mask RCNN & EE & RL & 0.590 & 0.221 \\
    InternImage & InternImage & EVE & RL & \textbf{0.702} & \textbf{0.252} \\
    \midrule
    SwinTrans & Mask RCNN & EVE & RL & \textbf{0.684} & \textbf{0.241}  \\
    SwinTrans & Mask RCNN & EVE & Frontier & 0.628 & 0.180 \\
    SwinTrans & Mask RCNN & EVE & Random & 0.636 & 0.236 \\
    SwinTrans & Mask RCNN & EE & RL & 0.620 & 0.216 \\
    SwinTrans & InternImage & EVE & RL & 0.676 & 0.238 \\
    \midrule
    GroundTruth & Mask RCNN & EVE & RL & 0.682 & 0.238 \\
    GroundTruth & Oracle & - & RL & 0.850 & 0.399 \\
    \bottomrule
  \end{tabular}
  }
  \vspace{-6pt}
  \caption{Variants of our method with different settings of instance classification, perception, switch policy and exploration policy. Perception means the Instance Segmentation model variations in Online Mapping module. ``EE" refers to the exploration-exploitation policy, which establishes a fixed threshold for decision-making. ``EVE" represents our Exploration-Verification-Exploitation Switch policy. The last row gives the upper bound of our method, where with ground truth instance segmented observations, the agent can identify the specified object instance as soon as the object becomes visible.}
  \vspace{-10pt}
  \label{tab:ablations}

\end{table}

%% file: sections/conclusion.tex
\section{Conclusion}

We propose an innovative framework of Instance-aware Exploration-Verification-Exploitation for Instance ImageGoal Navigation. We rethink previous work that decouple the decision-making process into an Exploration-Exploitation paradigm. By considering human behavior in the ``approach-verify" phase of identifying potential targets, we propose a new decision-making paradigm: Exploration-Verification-Exploitation. Our experimental results demonstrate that the Exploration-Verification-Exploitation paradigm effectively aids agents in their planning and decision-making process. Our proposed model significantly outperforms existing methods on the Instance ImageGoal Navigation task (0.684 \textit{vs.} 0.561 \textit{Success}) with a lightweight instance segmentation model. In the future, we will focus on enhancing the context of visual reasoning for more efficient exploration, and extend the application of the three-stage planning and decision-making paradigm to a broader range of tasks of embodied vision.